\pdfoutput=1
\documentclass[11pt,a4paper]{article}
\usepackage[hyperref]{acl2021}
\usepackage{times}
\usepackage{latexsym}

\usepackage{microtype}

\aclfinalcopy 
\def\aclpaperid{723} 

\usepackage{amsmath,amsfonts,bm}

\def\eqref#1{equation~\ref{#1}}

\def\1{\bm{1}}

\def\vu{{\bm{u}}}

\def\vx{{\bm{x}}}
\def\vy{{\bm{y}}}

\DeclareMathAlphabet{\mathsfit}{\encodingdefault}{\sfdefault}{m}{sl}
\SetMathAlphabet{\mathsfit}{bold}{\encodingdefault}{\sfdefault}{bx}{n}

\def\sU{{\mathbb{U}}}

\def\sY{{\mathbb{Y}}}

\newcommand{\mcZ}{\mathcal{Z}}
\newcommand{\mcL}{\mathcal{L}}
\newcommand{\mcdp}{\text{DP}}

\newcommand{\Score}{\text{Score}}

\newcommand{\F}{\text{F}}

\newcommand{\Se}{\text{S}_\text{e}}
\newcommand{\St}{\text{S}_\text{t}}
\usepackage{caption, subcaption}
\usepackage{graphicx}

\usepackage{amsmath}
\usepackage{amsfonts}
\usepackage{amssymb}

\usepackage{tikz,pgfplots}
\usepackage{tikz-dependency}

\usepackage{multicol, multirow}
\usepackage{hhline}
\usepackage{boldline}
\usepackage{mathtools}
\usepackage{enumitem}
\usepackage{hyperref}
\DeclareMathOperator*{\ssum}{\sum\sum}
\pgfplotsset{compat=1.17}

\title{Structural Knowledge Distillation: Tractably Distilling Information \\ for Structured Predictor}

\author{Xinyu Wang$^{\diamond\ddagger\spadesuit}$, Yong Jiang$^{\dagger}$\textsuperscript{$\ast$}$^{\spadesuit}$, Zhaohui Yan$^{\diamond\spadesuit}$, Zixia Jia$^{\diamond\spadesuit}$, Nguyen Bach$^{\dagger}$, Tao Wang$^{\dagger}$,\\
\textbf{Zhongqiang Huang$^{\dagger}$, Fei Huang$^{\dagger}$,  Kewei Tu$^{\diamond}$}\thanks{\hspace{1mm} Yong Jiang and Kewei Tu are the corresponding authors. $^{\spadesuit}$: Equal contributions. $^{\ddagger}$: This work was conducted when Xinyu Wang was interning at Alibaba DAMO Academy.} \\
 $^\diamond$School of Information Science and Technology, ShanghaiTech University \\
 Shanghai Engineering Research Center of Intelligent Vision and Imaging \\
 Shanghai Institute of Microsystem and Information Technology, Chinese Academy of Sciences \\
 University of Chinese Academy of Sciences \\
 $^\dagger$DAMO Academy, Alibaba Group \\
  {\tt \{wangxy1,jiazx,yanzhh,tukw\}@shanghaitech.edu.cn} \\
  {\tt \{yongjiang.jy,nguyen.bach\}@alibaba-inc.com} \\
  {\tt \{leeo.wangt,z.huang,f.huang\}@alibaba-inc.com} \\
}
\date{}

\begin{document}
\maketitle
\begin{abstract}
Knowledge distillation is a critical technique to transfer knowledge between models, typically from a large model (the teacher) to a more fine-grained one (the student). The objective function of knowledge distillation is typically the cross-entropy between the teacher and the student's output distributions. However, for structured prediction problems, the output space is exponential in size; therefore, the cross-entropy objective becomes intractable to compute and optimize directly. In this paper, we derive a factorized form of the knowledge distillation objective for structured prediction, which is tractable for many typical choices of the teacher and student models. In particular, we show the tractability and empirical effectiveness of structural knowledge distillation between sequence labeling and dependency parsing models under four different scenarios: 1) the teacher and student share the same factorization form of the output structure scoring function; 2) the student factorization produces more fine-grained substructures than the teacher factorization; 3) the teacher factorization produces more fine-grained substructures than the student factorization; 4) the factorization forms from the teacher and the student are incompatible.\footnote{Our code is publicly available at \url{https://github.com/Alibaba-NLP/StructuralKD}.}
\end{abstract}

\section{Introduction}
\label{sec:intro}
Deeper and larger neural networks have led to significant improvement in accuracy in various tasks, but they are also more computationally expensive and unfit for resource-constrained scenarios such as online serving. An interesting and viable solution to this problem is knowledge distillation (KD) \citep{Bucilua:2006:MC:1150402.1150464,NIPS2014_5484,44873}, which can be used to transfer the knowledge of a large model (the teacher) to a smaller model (the student). 
In the field of natural language processing (NLP), for example, KD has been successfully applied to compress massive pretrained language models such as BERT \citep{devlin-etal-2019-bert} and XLM-R \citep{conneau-etal-2020-unsupervised} into much smaller and faster models without significant loss in accuracy \citep{tang2019distilling,sanh2019distilbert,tsai-etal-2019-small,mukherjee-hassan-awadallah-2020-xtremedistil}.

A typical approach to KD is letting the student mimic the teacher model's output probability distributions on the training data by using the cross-entropy objective.
For structured prediction problems, however, the output space is exponentially large, making the cross-entropy objective intractable to compute and optimize directly.
Take sequence labeling for example. If the size of the label set is $L$, then there are $L^n$ possible label sequences for a sentence of $n$ words and it is infeasible to compute the cross-entropy by enumerating the label sequences. 
Previous approaches to structural KD either choose to perform KD on local decisions or substructures instead of on the full output structure, or resort to Top-K approximation of the objective \citep{kim-rush-2016-sequence,kuncoro-etal-2016-distilling,wang-etal-2020-structure}.

In this paper, we derive a factorized form of the structural KD objective based on the fact that almost all the structured prediction models factorize the scoring function of the output structure into scores of substructures. If the student's substructure space is polynomial in size and the teacher's marginal distributions over these substructures can be tractably estimated, then we can tractably compute and optimize the factorized form of the structural KD objective. As will be shown in the paper, many widely used structured prediction models satisfy the assumptions and hence are amenable to tractable KD. In particular, we show the feasibility and empirical effectiveness of structural KD with different combinations of teacher and student models, including those with incompatible factorization forms. 
We apply this technique to structural KD between sequence labeling and dependency parsing models under four different scenarios.


\begin{enumerate} [leftmargin=*]
    \item The teacher and student share the same factorization form of the output structure scoring function.
    \item The student factorization produces more fine-grained substructures than the teacher factorization.
    \item The teacher factorization produces more fine-grained substructures than the student factorization.
    \item The factorization forms from the teacher and the student are incompatible.
\end{enumerate}

In all the cases, we empirically show that our structural KD approaches can improve the student models. In the few cases where previous KD approaches are applicable, we show our approaches outperform these previous approaches.
With unlabeled data, our approaches can further improve student models' performance. 
In a zero-shot cross-lingual transfer case, we show that with sufficient unlabeled data, student models trained by our approaches can even outperform the teacher models.






\section{Background}

\subsection{Structured Prediction}
\label{sec:str_pred}
Structured prediction aims to predict a structured output such as a sequence, a tree or a graph. In this paper, we focus on structured prediction problems with a discrete output space, which include most of the structured prediction tasks in NLP (e.g., chunking, named entity recognition, and dependency parsing) and many structured prediction tasks in computer vision (e.g., image segmentation). We further assume that the scoring function of the output structure can be factorized into scores of a polynomial number of substructures. Consequently, we can calculate the conditional probability of the output structure $\vy$ given an input $\vx$ as follows:
\begin{align}
P(\vy|\vx) & = \frac{\exp{(\Score(\vy,\vx)})}{\sum_{\vy^{\prime} \in \sY(\vx)} \exp{(\Score(\vy^{\prime},\vx)})}\nonumber \\
            & = \frac{\prod_{\vu \in \vy}\exp{(\Score(\vu,\vx)})}{\mcZ(\vx)} 
\label{eq:2.1}
\end{align}
where $\sY(\vx)$ represents all possible output structures given the input $\vx$, $\Score(\vy,\vx)$ is the scoring function that evaluates the quality of the output $\vy$, $\mcZ(\vx)$ is the partition function, 
and $\vu \in \vy$ denotes that $\vu$ is a substructure of $\vy$. We define the substructure space
$\sU(\vx) = \bigcup_{\vy\in\sY(\vx)}\{\vu|\vu\in\vy \} $as the set of substructures of  all possible output structures given input $\vx$.

Take sequence labeling for example. Given a sentence $\vx$ , the output space $\sY(\vx)$ contains all possible label sequences of $\vx$. In linear-chain CRF, a popular model for sequence labeling, the scoring function $\Score(\vy,\vx)$ is computed by summing up all the transition scores
and emission scores 
where $i$ ranges over all the positions in sentence $\vx$, and the substructure space $\sU(\vx)$ contains all possible position-specific labels $\{y_i\}$ and label pairs $\{(y_{i-1},y_i)\}$.




\subsection{Knowledge Distillation}


\label{sec:kd}
Knowledge distillation is a technique that trains a small student model by encouraging it to imitate the output probability distribution of a large teacher model. The typical KD objective function is the cross-entropy between the output distributions predicted by the teacher model and the student model:
\begin{align}
\mcL_{\text{KD}} 
 = -\sum_{\vy \in \sY(\vx)} P_t(\vy|\vx)\log P_s(\vy|\vx) \label{eq:kd}
\end{align}
where $P_t$ and $P_s$ are the teacher's and the student's distributions respectively.

During training, the student jointly learns from the gold targets and the distributions predicted by the teacher by optimizing the following objective function:
\begin{align}
    \mcL_{\text{student}} = \lambda \mcL_\text{KD} + (1-\lambda) \mcL_\text{target} \nonumber
\end{align}
where $\lambda$ is an interpolation coefficient between the target loss $\mcL_\text{target}$ and the structural KD loss $\mcL_\text{KD}$. Following \citet{clark-etal-2019-bam,wang-etal-2020-structure}, one may apply teacher annealing in training by decreasing $\lambda$ linearly from 1 to 0. 
Because KD does not require gold labels, unlabeled data can also be used in the KD loss. 



\section{Structural Knowledge Distillation}
\label{sec:model}
When performing knowledge distillation on structured prediction, a major challenge is that the structured output space is exponential in size, leading to intractable computation of the KD objective in Eq. \ref{eq:kd}.
However, if the scoring function of the student model can be factorized into scores of substructures (Eq. \ref{eq:2.1}), then we can derive the following factorized form of the structural KD objective.
\begin{align}
&\mcL_{\text{KD}} 
= -\sum_{\vy \in \sY(\vx)} P_t(\vy|\vx){\log} P_s(\vy|\vx) \nonumber \\
& {=} {-}\smashoperator{\sum_{\vy \in \sY(\vx)}} P_t(\vy|\vx) \sum_{\vu \in \vy} \Score_s(\vu,\vx) {+} {\log} \mcZ_s(\vx) \nonumber\\
&{=} {-} \smashoperator{\sum_{\vy \in \sY(\vx)}} P_t(\vy|\vx) \smashoperator{\sum_{\vu \in \sU_s(\vx)}}  \1_{\vu \in \vy} \Score_s(\vu,\vx) {+} {\log} \mcZ_s(\vx) \nonumber \\
& {=} {-}\smashoperator{\ssum_{\substack{\vu \in \sU_s(\vx), \vy \in \sY(\vx)}}} P_t(\vy|\vx)  \1_{\vu \in \vy} \Score_s(\vu,\vx)  {+} {\log} \mcZ_s(\vx) \nonumber \\
& {=} {-}\smashoperator{\sum_{\vu \in \sU_s(\vx)}}P_t(\vu|\vx) \Score_s(\vu,\vx){+}{\log} \mcZ_s(\vx)
\label{eq:kl}
\end{align}
where $\1_\mathrm{condition}$ is 1 if the condition is true and 0 otherwise. From Eq. \ref{eq:kl}, we see that if 
$\sU_s(\vx)$ is polynomial in size and $P_t(\vu|\vx)$ can be tractably estimated, then the structural KD objective can be tractably computed and optimized.
In the rest of this section, we will show that this is indeed the case for some of the most widely used models in sequence labeling and dependency parsing, two representative structured prediction tasks in NLP. Based on the difference in score factorization between the teacher and student models, we divide our discussion into four scenarios.

\subsection{Teacher and Student Share the Same Factorization Form}

\paragraph{Case 1a: Linear-Chain CRF $\Rightarrow$ Linear-Chain CRF}
In this case, both the teacher and the student are linear-chain CRF models. 
An example application is to compress a state-of-the-art CRF model for named entity recognition (NER) that is based on large pretrained contextualized embeddings to a smaller CRF model with static embeddings that is more suitable for fast online serving.

For a CRF student model described in section \ref{sec:str_pred}, if we absorb the emission score $\Se(y_i,\vx)$ into the transition score $\St((y_{i-1},y_i),\vx)$ at each position $i$, then the substructure space $\sU_s(\vx)$ contains every two adjacent labels $\{(y_{i-1},y_i)\}$ for $i{=}1,\ldots,n$, with $n$ being the sequence length, and the substructure score is defined as $\Score((y_{i-1},y_i),\vx)= \St((y_{i-1},y_i),\vx)+\Se(y_i,\vx)$. The substructure marginal $P_t((y_{i-1}, y_i)|\vx)$ of the teacher model can be computed by:
\begin{align}
     P_t((y_{i-1}, y_i)|\vx) &\propto \alpha(y_{i-1}) \times \beta(y_i)\nonumber\\
     &\times \exp(\Score((y_{i-1},y_i),\vx))
     \label{eq:crf2crf}
\end{align}
where $\alpha(y_{i-1})$ and $\beta(y_i)$ are forward and backward scores that can be tractably calculated using the classical forward-backward algorithm. 

Comparing with the Posterior KD and Top-K KD of linear-chain CRFs proposed by \citet{wang-etal-2020-structure}, our approach calculates and optimizes the KD objective exactly, while their two KD approaches perform KD either heuristically or approximately. At the formulation level, our approach is based on the marginal distributions of two adjacent labels, while the Posterior KD is based on the marginal distributions of a single label.

\paragraph{Case 1b: Graph-based Dependency Parsing $\Rightarrow$ Dependency Parsing as Sequence Labeling}
In this case, we use the biaffine parser proposed by \citet{dozat-etal-2017-stanfords} as the teacher and the sequence labeling approach proposed by \citet{strzyz-etal-2019-viable} as the student for the dependency parsing task. 
The biaffine parser is one of the state-of-the-art models, while the sequence labeling parser provides a good speed-accuracy tradeoff. There is a big gap in accuracy between the two models and therefore KD can be used to improve the accuracy of the sequence labeling parser. 

Here we follow the head-selection formulation of dependency parsing without the tree constraint. The dependency parse tree $\vy$ is represented by $\langle y_1, \ldots, y_n \rangle$, where $n$ is the sentence length and $y_i=(h_i,l_i)$ denotes the dependency head of the $i$-th token of the input sentence, with $h_i$ being the index of the head token and $l_i$ being the dependency label.
The biaffine parser predicts the dependency head for each token independently. It models separately the probability distribution of the head index $P_t(h_i|\vx)$ and the probability distribution of the label $P_t( l_i |\vx)$. 
The sequence labeling parser is a MaxEnt model that also predicts the head of each token independently. 
It computes $\Score((h_i, l_i), \vx)$ for each token and applies a softmax function to produce the distribution $P_s((h_i,l_i)|\vx)$.

Therefore, these two models share the same factorization in which each substructure is a dependency arc specified by $y_i$. $\sU_s(\vx)$ thus contains all possible dependency arcs among tokens of the input sentence $\vx$. The substructure marginal predicted by the teacher can be easily derived as:
\begin{equation}
    P_t((h_i,l_i)|\vx) = P_t(h_i|\vx) \times P_t(l_i|\vx) 
    \label{eq:dp2sl}
\end{equation}
Note that in this case, the sequence labeling parser uses a MaxEnt decoder, which is locally normalized for each substructure. Therefore, the structural KD objective in Eq. \ref{eq:kl} can be reduced to the following form without the need for calculating the student partition function $\mcZ_s(\vx)$.
\begin{align}
    \mcL_{\text{KD}} 
    &= -\sum_{\vu \in \sU_s(\vx)}P_t(\vu|\vx) \times \text{log} P_s(\vu|\vx) 
\end{align}
In all the cases except \textbf{Case 1a} and \textbf{Case 3}, 
the student model is locally normalized and hence we can follow this form of objective.

\subsection{Student Factorization Produces More Fine-grained Substructures than Teacher Factorization}

\paragraph{Case 2a: Linear-Chain CRF $\Rightarrow$ MaxEnt}
\label{subsubsec:crf2soft}
In this case, we use a linear-chain CRF model as the teacher and a MaxEnt model as the student. Previous work \citep{yang-etal-2018-design,wang-etal-2020-structure} shows that a linear-chain CRF decoder often leads to better performance than a MaxEnt decoder for many sequence labeling tasks. Still, the simplicity and efficiency of the MaxEnt model is desirable. Therefore, it makes sense to perform KD from a linear-chain CRF to a MaxEnt model.
 
As mentioned in \textbf{Case 1a}, the substructures of a linear-chain CRF model are consecutive labels $\{(y_{i-1},y_i)\}$. In contrast, a MaxEnt model predicts the label probability distribution $P_s(y_i|\vx)$ of each token independently and 
hence the substructure space $\sU_s(\vx)$ consists of every individual label $\{y_i\}$.
To calculate the substructure marginal of the teacher $P_t(y_i|\vx)$, we can again utilize the forward-backward algorithm:
\begin{align}
    P_t(y_i|\vx) \propto \alpha(y_i) \times \beta(y_i) 
\end{align}
where $\alpha(y_{i})$ and $\beta(y_i)$ are forward and backward scores.


\paragraph{Case 2b: Second-Order Dependency Parsing $\Rightarrow$ Dependency Parsing as Sequence Labeling}

The biaffine parser is a first-order dependency parser, which scores each dependency arc in a parse tree independently. A second-order dependency parser scores pairs of dependency arcs with a shared token. The substructures of second-order parsing are therefore all the dependency arc pairs with a shared token.
It has been found that second-order extensions of the biaffine parser often have higher parsing accuracy \citep{wang-etal-2019-second,zhang-etal-2020-efficient,wang-etal-2020-enhanced,wang-tu-2020-second}.
Therefore, we may take a second-order dependency parser as the teacher to improve a sequence labeling parser.

Here we consider the second-order dependency parser of \citet{wang-tu-2020-second}. It employs mean field variational inference to estimate the probabilities of arc existence $P_t(h_i|\vx)$ and uses a first-order biaffine model to estimate the probabilities of arc labels $P_t(l_i|\vx)$. 
Therefore, the substructure marginal can be calculated in the same way as Eq. \ref{eq:dp2sl}.

\subsection{Teacher Factorization Produces More Fine-grained Substructures than Student Factorization}
\label{subsec:soft2crf}
\paragraph{Case 3: MaxEnt $\Rightarrow$ Linear-Chain CRF}
Here we consider KD in the opposite direction of \textbf{Case 2a}. An example application is zero-shot cross-lingual NER. 
Previous work \citep{pires-etal-2019-multilingual,wu-dredze-2019-beto} has shown that multilingual BERT (M-BERT) has strong zero-shot cross-lingual transferability in NER tasks. Many such models employ a MaxEnt decoder. 
In scenarios requiring fast speed and low computation cost, however, we may want to distill knowledge from such models to a model with much cheaper static monolingual embeddings while compensating the performance loss with a linear-chain CRF decoder. 

As described in \textbf{Case 1a}, the substructures of a linear-chain CRF model are consecutive labels $\{(y_{i-1},y_i)\}$. Because of the label independence and local normalization in the MaxEnt model, the substructure marginal of the MaxEnt teacher is calculated by:
\begin{align}
    P_t((y_{i-1}, y_i)|\vx) = P_t(y_{i-1}|\vx)P_t(y_i|\vx) 
    \label{max2crf}
\end{align}

\subsection{Factorization Forms From Teacher and Student are Incompatible}
\paragraph{Case 4: NER as Parsing $\Rightarrow$ MaxEnt}
Very recently, \citet{yu-etal-2020-named} propose to solve the NER task as graph-based dependency parsing and achieve state-of-the-art performance. They represent each named entity with a dependency arc from the first token to the last token of the named entity, and represent the entity type with the arc label. However, for the flat NER task (i.e., there is no overlapping between entity spans), the time complexity of this method is higher than commonly used sequence labeling NER methods. In this case, we take a parsing-based NER model as our teacher and a MaxEnt model with the BIOES label scheme as our student.

The two models adopt very different representations of NER output structures. The parsing-based teacher model represents an NER output of a sentence with a set of labeled dependency arcs and defines its score as the sum of arc scores. The MaxEnt model represents an NER output of a sentence with a sequence of BIOES labels and defines its score as the sum of token-wise label scores. Therefore, the factorization forms of these two models are incompatible.

Computing the substructure marginal of the teacher $P_t(y_i|\vx)$, where $y_i \in \{B_l, I_l, E_l, S_l, O | l \in L\}$ and $L$ is the set of entity types, is much more complicated than in the previous cases. Take $y_i = B_l$ for example. $P_t(y_i = B_l |\vx)$ represents the probability of the $i$-th word being the beginning of a multi-word entity of type `$l$’. In the parsing-based teacher model, this probability is proportional to the summation of exponentiated scores of all the output structures that contain a dependency arc of label `$l$’ with the $i$-th word as its head and with its length larger than 1. It is intractable to compute such marginal probabilities by enumerating all the output structures, but 
we can tractably compute them using dynamic programming. See supplementary material for a detailed description of our dynamic programming method.



\section{Experiments}
\label{sec:exp}
We evaluate our approaches described in Section \ref{sec:model} on NER (\textbf {Case 1a, 2a, 3, 4}) and dependency parsing (\textbf{Case 1b, 2b}). 

\subsection{Settings}
\label{sec:set}
\paragraph{Datasets} We use CoNLL 2002/2003 datasets \citep{tjong-kim-sang-2002-introduction,tjong-kim-sang-de-meulder-2003-introduction} for {\bf Case 1a}, \textbf{2a} and \textbf{4}, and use WikiAnn datasets \citep{pan-etal-2017-cross} for  {\bf Case 1a}, \textbf{2a}, \textbf{3}, and \textbf{4}. The CoNLL datasets contain the corpora of four Indo-European languages. We use the same four languages from the WikiAnn datasets. For cross-lingual transfer in \textbf{Case 3}, we use the four Indo-European languages as the source for the teacher model and additionally select four languages from different language families as the target for the student models.\footnote{The four languages from the CoNLL datasets are Dutch, English, German and Spanish and the four target languages for \textbf{Case 3} are Basque, Hebrew, Persian and Tamil. We use ISO 639-1 language codes (\url{https://en.wikipedia.org/wiki/List_of_ISO_639-1_codes}) to represent each language.} 

We use the standard training/development/test split for the CoNLL datasets. For WikiAnn, we follow the sampling of \citet{wang-etal-2020-structure} with 12000 sentences for English and 5000 sentences for each of the other languages. We split the datasets by 3:1:1 for training/development/test. For \textbf{Case 1b} and \textbf{2b}, we use Penn Treebank (PTB) 3.0 and follow the same pre-processing pipeline as in \citet{ma-etal-2018-stack}. 
For unlabeled data, we sample sentences that belong to the same languages of the labeled data from the WikiAnn datasets for {\bf Case 1a}, \textbf{2a} and {\bf 4} and we sample sentences from the target languages of WikiAnn datasets for \textbf{Case 3}. We use the BLLIP corpus\footnote{Brown Laboratory for Linguistic Information Processing (BLLIP) 1987-89 WSJ Corpus Release 1. } as the unlabeled data for \textbf{Case 1b} and \textbf{2b}.

\paragraph{Models}

For the student models in all the cases, we use fastText \citep{bojanowski2017enriching} word embeddings and character embeddings as the word representation. For {\bf Case 1a}, \textbf{2a} and {\bf 4}, we concatenate the multilingual BERT, Flair \citep{akbik-etal-2018-contextual}, fastText embeddings and character embeddings \citep{santos2014learning} as the word representations for stronger monolingual teacher models \cite{wang-etal-2020-more}. For {\bf Case 3}, we use M-BERT embeddings for the teacher. Also for \textbf{Case 3}, we fine-tune the teacher model on the training set of the four Indo-European languages from the WikiAnn dataset and train student models on the four additional languages. For the teacher models in {\bf Case 1b} and {\bf 2b}, we simply use the same embeddings as the student because there is already huge performance gap between the teacher and student in these settings and hence we do not need strong embeddings for the teacher to demonstrate the utility of KD.

\paragraph{Baselines} 
We compare our Structural KD (\textit{Struct. KD}) with training without KD (\textit{w/o KD}) as well as existing KD approaches.
In \textbf{Case 1a}, the \textit{Pos. KD} baseline is the Posterior KD approach for linear-chain CRFs proposed by \citet{wang-etal-2020-structure}. They also propose \textit{Top-K KD} but have shown that it is inferior to \textit{Pos. KD}. For experiments using unlabeled data in all the cases, in addition to labeled data, we use the teacher's prediction on the unlabeled data as pseudo labeled data to train the student models. This can be seen as the \textit{Top-1 KD} method\footnote{We do not predict pseudo labels for the labeled data, because we find that the teacher models' predictions on the labeled training data have approximately 100\% accuracy in most of the cases. }. 
In Case \textbf{2a} and \textbf{3}, where we perform KD between CRF and MaxEnt models, we run a reference baseline that replaces the CRF teacher or student model with a MaxEnt model and performs token-level KD (\textit{Token KD}) of MaxEnt models that optimizes the cross entropy between the teacher and student label distributions at each position.


\begin{table}[t]
\small
\centering
\setlength\tabcolsep{1pt}
\begin{tabular}{l|cc|c|cc|c|cc}
\hlineB{4}

Case     & \multicolumn{2}{c|}{\textbf{1a}} & \textbf{1b} & \multicolumn{2}{c|}{\textbf{2a}} & \textbf{2b} & \multicolumn{2}{c}{\textbf{4}} \\

Labeled      & CoN            & Wiki            & PTB              & CoN            & Wiki            & PTB              & CoN           & Wiki           \\ \hline\hline
Teacher      & 89.15            & 88.52              & 95.96            & 89.15            & 88.52              & 96.04            & 88.57           & 88.38             \\
\hline
w/o KD        & 84.70            & 83.31              & 89.85            & 83.87            & 80.86              & 89.85            & 83.87           & 80.86             \\
Pos. KD & 85.27            & 83.73              & -                & -                & -                  & -                & -               & -                 \\
Struct. KD    & \textbf{85.35}   & \textbf{84.12}     & \textbf{91.83}   & \textbf{84.50}   & \textbf{82.23}     & \textbf{91.78}   & \textbf{84.28}  & \textbf{81.45}    \\ 
\hlineB{4}
\end{tabular}
\caption{Averaged F1 scores for NER and labeled attachment scores (LAS) for dependency parsing on labeled datasets. CoN: CoNLL datasets.}
\label{tab:labeled}
\end{table}

\begin{table}[t]
\small
\centering
\setlength\tabcolsep{1.5pt}
\begin{tabular}{l|c|c|c|c|c|c}
\hlineB{4}
Case       & \textbf{1a} & \textbf{1b} & \textbf{2a} & \textbf{2b} & \textbf{3} & \textbf{4} \\ 
Labeled+Unlabeled        & Wiki           & PTB           & Wiki           & PTB            & Wiki U          & Wiki          \\ 
\hline\hline
Teacher        & 88.52            & 95.96            & 88.52            & 96.04            & 56.01           & 88.38           \\
\hline
Top-1       & 84.19            & 90.03            & 82.40             & 90.03            & 41.11           & 82.10            \\
Pos. KD + Top-1   & 84.91            & -                & -                & -                & -               & -               \\
Struct. KD + Top-1 & \textbf{85.24}   & \textbf{91.98}   & \textbf{85.24}   & \textbf{91.94}   & \textbf{45.28}  & \textbf{82.44}  \\ \hlineB{4}
\end{tabular}
\caption{Average F1 score of NER and labeled attachment scores (LAS) for dependency parsing with both labeled and unlabeled data. Wiki U means that the training data of this case contains only the unlabeled data. }
\label{tab:unlabel}
\end{table}

\paragraph{Training}
For MaxEnt and linear-chain CRF models, we use the same hyper-parameters as in \citet{akbik-etal-2018-contextual}. For dependency parsing, we use the same hyper-parameters as in \citet{wang-tu-2020-second} for teacher models and \citet{strzyz-etal-2019-viable} for student models. For M-BERT fine-tuning in {\bf Case 3}, we mix the training data of the four source datasets and train the teacher model with the AdamW optimizer \citep{loshchilov2018decoupled} with a learning rate of $5{\times} 10^{-5}$ for 10 epochs. We tune the KD temperature in $\{1,2,3,4,5\}$ and the loss interpolation annealing rate in $\{0.5,1.0,1.5\}$. 
For all experiments, we train the models for 5 runs with a fixed random seed for each run.

\subsection{Results}
\label{sec:results}

\begin{table}[t]
\small
\centering
\setlength\tabcolsep{2pt}
\begin{tabular}{l||ccc|c}
\hlineB{4}
                 & \multicolumn{3}{c|}{\textbf{Case 2a}}                    & \textbf{Case 3}         \\
                 & CoNLL         & WikiAnn        & Wiki+U         & Wiki U         \\ \hline\hline
MaxEnt Teacher & 88.65         & 87.41          & 87.41          & 56.01          \\
CRF Teacher      & 89.15         & 88.52          & 88.52          & -            \\
\hline
Token. KD        & 84.25         & 82.09          & 83.07          & 38.42          \\
Struct. KD       & \textbf{84.50} & \textbf{82.23} & \textbf{83.34} & \textbf{45.28} \\ \hlineB{4}
\end{tabular}
\caption{Comparing with reference baselines on NER task. Wiki+U means the training data comprises labeled and unlabeled WikiAnn data and Wiki U means that the training data of this case contains only the unlabeled data.}
\label{tab:baseline_ref}
\end{table}

Table \ref{tab:labeled} shows the experimental results with labeled data only and \ref{tab:unlabel} shows the experimental results with additional 3000 unlabeled sentences.
The results show that our structural KD approaches outperform the baselines in all the cases. Table \ref{tab:baseline_ref} compares \textit{Struct. KD} with \textit{Token KD}, the reference baseline based on MaxEnt models.
For \textbf{Case 2a}, which involves a MaxEnt student, \textit{Struct. KD} with a CRF teacher achieves better results than \textit{Token KD} with a MaxEnt teacher. For \textbf{Case 3}, which involves a MaxEnt teacher, \textit{Struct. KD} with a CRF student achieves better results than \textit{Token KD} with a MaxEnt student. 
These results are to be expected because \textit{Struct. KD} makes it possible to apply exact knowledge distillation with a more capable teacher or student. 
In all the experiments, we run Almost Stochastic Dominance proposed by \citet{dror-etal-2019-deep} with a significance level of $0.05$ and find that the advantages of our structural KD approaches are significant. Please refer to Appendix for more detailed results.

\begin{table}[t]
\setlength\tabcolsep{5pt}
\centering
\small
\begin{tabular}{l||cccc|c}
\hlineB{4}
&	 de    	 &   en &	  es    	 & nl    &	  Avg.  \\
\hline					
\multicolumn{6}{c}{\textbf{Case 1a}}\\					
\hline					
w/o KD$^\dagger$         &      	82.16&	90.13&	88.06&	89.11&	 87.36\\
Top-WK KD$^\dagger$  &    	82.15&	90.52&	88.64&	89.24&	   87.64\\
Pos. KD$^\dagger$  &            	82.22&	90.68&	88.57&	89.41&	 87.72\\
Pos.+Top-WK$^\dagger$ &	 \textbf{82.31} &	90.53	&88.66&	89.58&	 87.77\\
Struct. KD& 	82.28&	 \textbf{90.86}& 	 \textbf{88.67}& 	 \textbf{90.07}& 	 \textbf{87.97} \\
\hline
\multicolumn{6}{c}{\textbf{Case 2a}}\\
\hline
w/o KD$^\dagger$ &	 \textbf{81.40}  &	90.08 &	87.72 &	88.99	 & 87.05 \\
Token KD$^\dagger$  &	81.30 &	90.02 &	88.24 &	88.87 &	 87.11 \\
Struct. KD  &	81.27 &	 \textbf{90.25}  &	 \textbf{88.64}  &	 \textbf{89.14}  &	 \textbf{87.32}\\
\hlineB{4}
\end{tabular}
\caption{A comparison of KD approaches for multilingual NER. $\dagger$: Results are from \citet{wang-etal-2020-structure}.}
\label{tab:ner}
\end{table}

\subsection{Multilingual NER Experiments}
There is a recent increase of interest in training multilingual NER models \citep{tsai-etal-2019-small,mukherjee-hassan-awadallah-2020-xtremedistil} because of the strong generalizability of M-BERT on multiple languages. Existing work explored knowledge distillation approaches to train fast and effective multilingual NER models with the help of monolingual teachers \citep{wang-etal-2020-structure}. To show the effectiveness of structural KD in the multilingual NER setting, we compare our approaches with those reported by \citet{wang-etal-2020-structure}.
Specifically, the monolingual teachers are always CRF models, and the multilingual student is either a CRF model (\textbf{Case 1a}) or a MaxEnt model (\textbf{Case 2a}). \citet{wang-etal-2020-structure} report results of the \textit{Top-WK KD} (a weighted version of \textit{Top-K KD}) and \textit{Pos. KD} approaches for \textbf{Case 1a} and the reference baseline \textit{Token KD} (with a MaxEnt teacher) for \textbf{Case 2a}. We follow their experimental settings when running our approach.


The experimental results in Table \ref{tab:ner} show the effectiveness of \textit{Struct. KD} in both cases. 
In \textbf{Case 1a}, our approach is stronger than both \textit{Top-WK KD} and \textit{Pos. KD} as well as the mixture of the two approaches on average. 
In \textbf{Case 2a}, \textit{Struct. KD} 
not only outperforms \textit{Token KD}, but also makes the MaxEnt student competitive with the CRF student without KD (87.32 vs. 87.36).  



\begin{figure*}[!ht]
\centering
\begin{tikzpicture}
    \node at (-4.7,-0.65) {(a) Case 1a};
    \node at (0.7,-0.65) {(b) Case 1b};
    \node at (5.9,-0.65) {(c) Case 3};
    \begin{axis}[
        xshift=-7cm,
        name=ner,
        width=0.38\textwidth,
        height=0.25\textwidth,
        legend columns=2, 
        legend pos=north west,
        legend style={font=\tiny,yshift=-0.3cm},
        tick label style={font=\small},
        xtick={0,3,10,30},
        ylabel style={yshift=-0.5cm},
        ]
        \addplot[red,mark=square] table[x=unlabel,y=crf] {crf2crf.dat};
        \addplot[black,mark=*] table[x=unlabel,y=posterior] {crf2crf.dat};
        \addplot[blue,mark=square*] table[x=unlabel,y=structural] {crf2crf.dat};
        \addplot[black,dashed,line width=0.4mm] table[x=unlabel,y=teacher] {crf2crf.dat};
        \addplot[black,dotted,line width=0.4mm] table[x=unlabel,y=baseline] {crf2crf.dat};
        \legend{Top-1 KD, Pos. KD, Struct. KD}
    \end{axis}
    \begin{axis}[
        xshift=5.3cm,
        at={(ner.south west)},
        name=zeroshot,
        width=0.38\textwidth,
        height=0.25\textwidth,
        legend columns=1, 
        legend pos=south east,
        legend style={font=\tiny,yshift=0.3cm},
        tick label style={font=\small},
         xtick={0,10,30,50,100},
        ylabel style={yshift=-0.5cm},
        ]
        \addplot[red,mark=square] table[x=unlabel,y=Baseline] {graph2seq_parsing.dat};
        \addplot[blue,mark=square*] table[x=unlabel,y=Structural] {graph2seq_parsing.dat};
        \addplot[black,dashed,line width=0.4mm] table[x=unlabel,y=Teacher] {graph2seq_parsing.dat};
        \addplot[black,dotted,line width=0.4mm] table[x=unlabel,y=baseline] {graph2seq_parsing.dat};
        \legend{Top-1 KD, Struct. KD}
    \end{axis}
    \begin{axis}[
        at={(zeroshot.south west)},
        xshift=5.3cm,
        width=0.38\textwidth,
        height=0.25\textwidth,
        legend columns=1, 
        legend pos=south east,
        legend style={font=\tiny},
        tick label style={font=\small},
        xtick={0, 3,10,30},
        ylabel style={yshift=-0.5cm},
        ]
        \addplot[red,mark=square] table[x=unlabel,y=CRF] {zs_maxent2crf.dat};
        \addplot[black,mark=*] table[x=unlabel,y=MaxEnt->MaxEnt] {zs_maxent2crf.dat};
        \addplot[blue,mark=square*] table[x=unlabel,y=Structural] {zs_maxent2crf.dat};
        \addplot[black,dashed,line width=0.4mm] table[x=unlabel,y=Teacher] {zs_maxent2crf.dat};
        \legend{Top-1 KD, Token KD, Struct. KD}
       
    \end{axis}
\end{tikzpicture}
\caption{The accuracy of structural KD and the baselines on different amounts of unlabeled data in three cases. The x-axis represents the amount of unlabeled data in thousand and the y-axis represents the accuracy. The dashed lines are the accuracy of the teacher models. The dotted lines are the accuracy of the baseline models without any knowledge from the teachers.}
\label{fig:performance_curve}
\end{figure*}
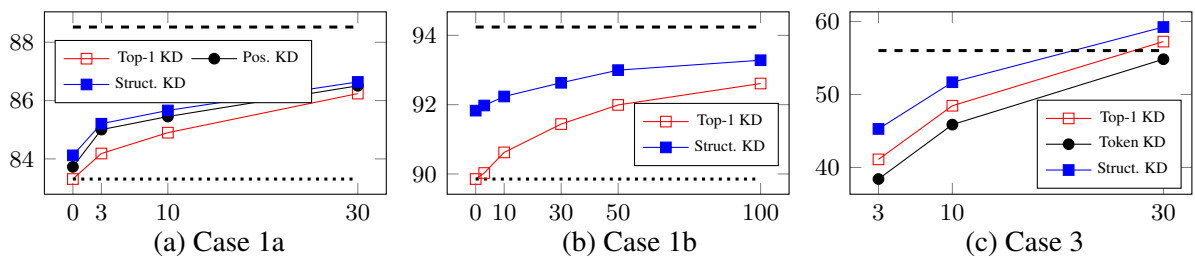

\section{Analysis}
\subsection{Amount of Unlabeled Data}

We compare our approaches with the baselines with different amounts of unlabeled data for \textbf{Case 1a}, \textbf{1b} and \textbf{3}, which are cases that apply in-domain unlabeled data for NER and dependency parsing, and cross-lingual unlabeled data for NER. We experiment with more unlabeled data for \textbf{Case 1b} than for the other two cases because the labeled training data of PTB is more than $10$ times larger than the labeled NER training data in \textbf{Case 1a} and \textbf{3}. Results are shown in Figure \ref{fig:performance_curve}. 
The experimental results show that our approaches consistently outperform the baselines, though the performance gaps between them become smaller when the amount of unlabeled data increases. Comparing the performance of the students with the teachers, we can see that in \textbf{Case 1a} and \textbf{1b}, the gap between the teacher and the student remains large even with the largest amount of unlabeled data. This is unsurprising considering the difference in model capacity between the teacher and the student. In \textbf{Case 3}, however, we find that when using 30,000 unlabeled sentences, the CRF student models can even outperform the MaxEnt teacher model, which shows the effectiveness of CRF models on NER. 


\subsection{Temperature in Structural Knowledge Distillation}
A frequently used KD technique is dividing the logits of probability distributions of both the teacher and the student by a temperature in the KD objective \citep{44873}. Using a higher temperature produces softer probability distributions and often results in higher KD accuracy. In structural KD, there are two approaches to applying the temperature to the teacher model, either globally to the logit of $P_t(\vy|\vx)$ (i.e., $\Score_t(\vy,\vx)$) of the full structure $\vy$, or locally to the logit of $P_t(\vu|\vx)$ of each student substructure $\vu$. 
We empirically compare these two approaches in \textbf{Case 1a} with the same setting as in Section \ref{sec:set}. Table \ref{tab:temperature} shows that the local approach results in better accuracy for all the languages. Therefore, we use the local approach by default in all the experiments.

\begin{table}[!t]

\small
\centering
\begin{tabular}{l||cccc|c}
\hlineB{4}
 & de    & en    & es    & nl    & Avg.   \\
\hline\hline
CRF    & 75.37 & 91.21 & 86.55 & 85.67 & 84.70 \\
Global      & 75.67 & 91.11 & 86.72 & 85.92 & 84.85 \\
Local       & \textbf{76.61} & \textbf{91.41} & \textbf{87.20} & \textbf{86.19} & \textbf{85.35}\\
\hlineB{4}
\end{tabular}
\captionof{table}{Comparison of the global and local temperature application approaches on CoNLL NER.}
\label{tab:temperature}
\end{table}
\begin{table}[!t]
\small \centering
\begin{tabular}{l||cc}
\hlineB{4}
                  & CoNLL   & WikiAnn \\
\hline\hline
CRF                & \textbf{89.15}  & \textbf{88.52} \\
CRF-Mrg.      & 89.08 & 88.41 \\
NER-Par.           & 88.57   & 88.38 \\
NER-Par.-Mrg. & 87.40   & 86.82 \\
MaxEnt            & 88.65 & 87.41  \\
\hlineB{4}
\end{tabular}
\captionof{table}{Averaged F1 score of teachers and it's marginal distributions. -Mrg.: Marginal distribution, NER-Par.: NER as parsing \citep{yu-etal-2020-named}.}
\label{tab:teachers}
\end{table}

\subsection{Comparison of Teachers}
In \textbf{Case 2a} and \textbf{Case 4}, we use the same MaxEnt student model but different types of teacher models. Our structural KD approaches in both cases compute the marginal distribution $P_t(y_i|\vx)$ of the teacher at each position $i$ following the substructures of the MaxEnt student, which is then used to train the student substructure scores. We can evaluate the quality of the marginal distributions by taking their modes as label predictions and evaluating their accuracy.
In Table \ref{tab:teachers}, we compare the accuracy of the CRF teacher and its marginal distributions from \textbf{Case 2a}, the NER-as-parsing teacher and its marginal distributions from \textbf{Case 4}, and the MaxEnt teacher which is the KD baseline in \textbf{Case 2a}.
First, we observe that for both CRF and NER-as-parsing, predicting labels from the marginal distributions leads to lower accuracy. This is to be expected because such predictions do not take into account correlations between adjacent labels. While predictions from marginal distributions of the CRF teacher still outperform MaxEnt, those of the NER-as-parsing teacher clearly underperform MaxEnt. This provides an explanation as to why \textit{Struct. KD} in \textbf{Case 4} has equal or even lower accuracy than the \textit{Token KD} baseline in \textbf{Case 2a} in Table \ref{tab:baseline_ref}.


\section{Related Work}

\subsection{Structured Prediction}

In this paper, we use sequence labeling and dependency parsing as two example structured prediction tasks.
In sequence labeling, a lot of work applied the linear-chain CRF and achieved state-of-the-art performance in various tasks \citep{ma-hovy-2016-end,akbik-etal-2018-contextual,liu-etal-2019-gcdt,yu-etal-2020-named,wei-etal-2020-dont,wang2020automated,wang2021improving}. Meanwhile, a lot of other work used the MaxEnt layer instead of the CRF for sequence labeling \citep{devlin-etal-2019-bert,conneau-etal-2020-unsupervised,wang-etal-2020-ain} because MaxEnt makes it easier to fine-tune pretrained contextual embeddings in training. Another advantage of MaxEnt in comparison with CRF is its speed. \citet{yang-etal-2018-design} showed that models equipped with the CRF are about two times slower than models with the MaxEnt layer in sequence labeling.
In dependency parsing, recent work shows that second-order CRF parsers achieve significantly higher accuracy than first-order parsers \citep{wang-etal-2019-second,zhang-etal-2020-efficient}. However, the inference speed of second-order parsers is much slower. \citet{zhang-etal-2020-efficient} showed that second-order parsing is four times slower than the simple head-selection first-order approach \citep{dozat2016deep}. 
Such speed-accuracy tradeoff as seen in sequence labeling and dependency parsing also occurs in many other structured prediction tasks. This makes KD an interesting and very useful technique that can be used to circumvent this tradeoff to some extent.


\subsection{Knowledge Distillation in Structured Prediction}
KD has been applied in many structured prediction tasks in the fields of NLP, speech recognition and computer vision, with applications such as neural machine translation \citep{kim-rush-2016-sequence,tan2018multilingual},  sequence labeling \citep{tu-gimpel-2019-benchmarking,wang-etal-2020-structure}, 
connectionist temporal classification \citep{Huang2018}, 
image semantic segmentation \citep{liu2019structured} and so on. 
In KD for structured prediction tasks, how to handle the exponential number of structured outputs is a main challenge. To address this difficult problem, recent work resorts to approximation of the KD objective. \citet{kim-rush-2016-sequence} proposed sequence-level distillation through predicting K-best sequences of the teacher in neural machine translation. \citet{kuncoro-etal-2016-distilling} proposed to use multiple greedy parsers as teachers and generate the probability distribution at each position through voting. Very recently, \citet{wang-etal-2020-structure} proposed structure-level knowledge distillation for linear-chain CRF models in multilingual sequence labeling. During the distillation process, teacher models predict the Top-K label sequences as the global structure information or the posterior label distribution at each position as the local structural information, which is then used to train the student. Besides approximate approaches, an alternative way is using models that make local decisions and performing KD on these local decisions. \citet{anderson-gomez-rodriguez-2020-distilling} formulated dependency parsing as a head-selection problem and distilled the distribution of the head node at each position. \citet{tsai-etal-2019-small} proposed MiniBERT through distilling the output distributions of M-BERT models of the MaxEnt classifier. Besides the output distribution, \citet{mukherjee-hassan-awadallah-2020-xtremedistil} further distilled the hidden representations of teachers.


\section{Conclusion}
In this paper, we propose structural knowledge distillation, which transfers knowledge between structured prediction models. We derive a factorized form of the structural KD objective and make it tractable to compute and optimize for many typical choices of teacher and student models. We apply our approach to four KD scenarios with six cases for sequence labeling and dependency parsing. Empirical results show that our approach outperforms baselines without KD as well as previous KD approaches. With sufficient unlabeled data, our approach can even boost the students to outperform the teachers in zero-shot cross-lingual transfer. 

\subsubsection*{Acknowledgments}
This work was supported by the National Natural Science Foundation of China (61976139) and by Alibaba Group through Alibaba Innovative Research Program. 

\bibliography{anthology,custom,acl2021}
\bibliographystyle{acl_natbib}

\clearpage
\appendix
\section{Dynamic Programming for Case 4}
\label{app:dp_alg}
We describe how the marginal distribution over BIOES labels at each position of the input sentence can be tractably computed based on the NER-as-parsing teacher model using dynamic programming.







Given an input sentence $\vx$ with $n$ words, we first define the following functions.
\begin{itemize}
    \item $\overrightarrow{\mcdp}(i,l)$ represents the summation of scores of all possible labeling sequences of the sub-sentence from the first token to the $i$-th token while a span ends with the $i$-th token with a label $l$.
    \item $\overrightarrow{\mcdp}(i,\F)$ represents the summation of scores of all possible labeling sequences of the sub-sentence from the first token to the $i$-th token while there is no arc pointing to the $i$-th token.
    \item $\overleftarrow{\mcdp}(i,l)$ represents the summation of scores of all possible labeling sequences of the sub-sentence from the $i$-th toke to the last token while a span starts with the $i$-th token with a label $l$.
    \item $\overleftarrow{\mcdp}(i,\F)$ represents the summation of scores of all possible labeling sequences of the sub-sentence from the $i$-th toke to the last token while there is no arc coming from the $i$-th token.
\end{itemize}
We can compute the values of these functions for all values of $i$ and $l$ using dynamic programming.
The base cases are:
\begin{gather*}
    \overrightarrow{\mcdp}(1,\F) = 1    \qquad\qquad \overleftarrow{\mcdp}(n,\F) = 1 \\
\end{gather*}
The recursive formulation of these functions are:
\begin{align}
    &\overrightarrow{\mcdp}(i,l) = \sum_{k=1}^i\exp(\Score(y_{k,i}=l)) * \overrightarrow{\mcdp}(k,\F)\nonumber\\
    &\overrightarrow{\mcdp}(i,\F) = \overrightarrow{\mcdp}(i-1,\F)+\sum_{l \in L}\overrightarrow{\mcdp}(i-1,l) \nonumber\\
    &\overleftarrow{\mcdp}(i,l) = \sum_{j=i}^n\exp(\Score(y_{i,j}=l)) * \overleftarrow{\mcdp}(j,\F)\nonumber\\
    &\overleftarrow{\mcdp}(i,\F) = \overleftarrow{\mcdp}(i+1,\F)+\sum_{l \in L}\overleftarrow{\mcdp}(i+1,l)\nonumber
\end{align}
where $\Score(y_{i,j}=l)$ is the score assigned by the teacher model to the dependency arc from $i$ to $j$ with label $l$.
After dynamic programming, we can compute the substructure marginals of the teacher $P_t(y_i|\vx)$ as follows:

\begin{align}
  & P_t(y_i=B_l|\vx) =\text{DP}(B_l, i)/\mcZ(\vx)\nonumber\\
    &=  \overrightarrow{\text{DP}}(i,\text{F}) * \sum_{j=i+1}^n \exp(\Score(y_{i,j}=l))\nonumber\\
    & \quad *\overleftarrow{\mcdp}(j,\text{F})/\mcZ(\vx)\nonumber \\
    \nonumber\\
    &P_t(y_i=I_l|\vx)
    = \mcdp(I_l,i)/\mcZ(\vx) \nonumber\\
    &=  \sum_{k=1}^{i-1}\sum_{j=i+1}^n\exp(\Score(y_{k,j}=l))*\overrightarrow{\mcdp}(k,\F) \nonumber\\
    &\quad *\overleftarrow{\mcdp}(j,\F)/\mcZ(\vx)\nonumber\\
  \nonumber \\
    &P_t(y_i=E_l|\vx) = \mcdp(E_l,i)/\mcZ(\vx)\nonumber\\
    &= \overleftarrow{\text{DP}}(i,\text{F}) * \sum_{k=1}^{i-1}\exp(\Score(y_{k,i}=l))\nonumber\\
    &\quad*\overrightarrow{\mcdp}(k,\text{F})/\mcZ(\vx) \nonumber\\
    \nonumber\\
    &P_t(y_i=O|\vx) = \mcdp(O,i)/\mcZ(\vx) \nonumber\\
    &= \overrightarrow{\mcdp}(i,\F) * \overleftarrow{\mcdp}(i,\F)/\mcZ(\vx) \nonumber \\
  \nonumber \\
    &P_t(y_i=S_l|\vx) = \mcdp(S_l,i)/\mcZ(\vx)\nonumber\\
    &= \overrightarrow{\mcdp}(i,\F) {*} \exp(\Score(y_{i,i}=l)) {*} \overleftarrow{\mcdp}(i,\F)/\mcZ(\vx) \nonumber
\end{align}
where
\begin{itemize}
    \item $\mcdp(X, i)$ represents the summation of scores of all possible labeling sequences in which the $i$-th token is labeled as $X$. $X$ can be one of `$B_l, I_l, E_l, O, S_l$'.
    \item $\mcZ(\vx)$ represents the summation of scores of all possible labeling sequences given the input sentence $\vx$. $y_{i,j}=l$ represents that there is a dependency arc of label ‘$l$’ from the $i$-th word to the $j$-th word. We can calculate $\mcZ(\vx)$ by
    $\overrightarrow{\mcdp}(n,l)+\overrightarrow{\mcdp}(n,F)$ or $\overleftarrow{\mcdp}(1,l)+\overleftarrow{\mcdp}(1,F)$
\end{itemize}
The edge cases are:
\begin{align}
    &P_t(y_n=B_l|\vx)=0  \nonumber\\   
    &P_t(y_1=I_l|\vx)=P_t(y_n=I_l|\vx)=0   \nonumber\\
    &P_t(y_1=E_l|\vx)=0\nonumber
\end{align}

\begin{table}[!t]
\small \centering
\begin{tabular}{l||cc}
\hlineB{4}
 & Speed (sentences/second)     & \# Param (M)   \\
\hline\hline
Teacher & 27.76 & 233.40 \\
Student & 672.20 &  9.46\\
\hlineB{4}
\end{tabular}
\captionof{table}{Running speed and model sizes of the teacher and student models in \textbf{Case 2a}.}
\label{tab:speed}
\end{table}

\section{Additional Analysis}

\subsection{Comparison of Speed and Model Size}

An important goal of KD is to produce faster and smaller models. In Table \ref{tab:speed}, we show a comparison on the running speed and model size between the teacher and student models on the CoNLL English test set from \textbf{Case 2a}. It can be seen that the student model is about 24 times faster and 25 times smaller than the teacher model.
\section{Detailed Experimental Results}
\label{app:results}
In this section, we present detailed experimental results.
Table \ref{tab:appendix_ner_labeled}, \ref{tab:appendix_ner_unlabel} and \ref{tab:appendix_zeroshot} show the results of NER task, while table \ref{tab:appendix_parsing_labeled} and \ref{tab:appendix_parsing_unlabeled} show the results of Parsing. We evaluate the significance based on Almost Stochastic Dominance (ASD) \citep{dror-etal-2019-deep}, which is a high quality comparison between deep neural networks. We evaluate with a significance level of $0.05$. For the significance test over averaged scores, we averaged over the same random seed of each language as a sample of averaged score.
In tables, we use $\dagger$ to represent our approaches are significantly stronger than the models training without KD or with Top-1 KD. We use $\ddagger$ to represent that our approaches are significantly stronger than other KD approaches.

\subsection{Results of NER task}
Table \ref{tab:appendix_ner_labeled}, \ref{tab:appendix_ner_unlabel} and \ref{tab:appendix_zeroshot} represent the KD results of experiments with labeled and unlabeled datasets. Our approaches outperform the baselines significantly in most of the cases. Note that in some cases, our approaches perform slightly inferior to other approaches (for example, \textbf{de} dataset in \textbf{Case 1a} in Table \ref{tab:appendix_ner_unlabel} with 30k unlabeled sentences) while our approaches are still stronger than these approaches according to the ASD test. The possible reason is that the variances of our approaches are much larger than the other approaches and ASD indicates our approaches is possibly better than the other approaches.

\begin{table*}[t!]
\small
\centering

\begin{tabular}{c|l|l|ccccc|ccccc}
\hlineB{4}
\multicolumn{3}{c|}{Dataset}                                      & \multicolumn{5}{c|}{CoNLL}                                                         & \multicolumn{5}{c}{WikiAnn}                                                       \\ \hline
\multicolumn{3}{c|}{Scenario}                                     & de             & en             & es             & nl             & Avg.           & de             & en             & es             & nl             & Avg.           \\ \hline\hline
\multicolumn{2}{c|}{\multirow{4}{*}{\textbf{Case 1a}}} & Teacher           & 83.48          & 92.25          & 89.29          & 91.56          & 89.15          & 86.98          & 83.80          & 91.85          & 91.46          & 88.52          \\ \cline{3-13} 
\multicolumn{2}{c|}{}                         & w/o KD          & 75.37          & 91.21          & 86.55          & 85.67          & 84.70          & 80.12          & 80.09          & 85.84          & 87.19          & 83.31          \\  
\multicolumn{2}{c|}{}                         & pos. KD      & 76.46          & 91.38          & \textbf{87.33} & 85.92          & 85.27          & 80.02          & \textbf{81.76} & 85.98          & 87.15          & 83.73          \\  
\multicolumn{2}{c|}{}                         & Struct. KD     & \textbf{76.61}\rlap{$^{\dagger\ddagger}$} & \textbf{91.41}\rlap{$^{\dagger\ddagger}$} & 87.20\rlap{$^{\dagger}$}          & \textbf{86.19}\rlap{$^{\dagger\ddagger}$} & \textbf{85.35}\rlap{$^{\dagger\ddagger}$} & \textbf{80.64}\rlap{$^{\dagger\ddagger}$} & 81.37\rlap{$^{\dagger}$}          & \textbf{87.29}\rlap{$^{\dagger\ddagger}$} & \textbf{87.19}\rlap{$^{\dagger\ddagger}$} & \textbf{84.12}\rlap{$^{\dagger\ddagger}$} \\ \hline
\multicolumn{2}{c|}{\multirow{5}{*}{\textbf{Case 2a}}} & CRF Teacher       & 83.48          & 92.25          & 89.29          & 91.56          & 89.15          & 86.98          & 83.80          & 91.85          & 91.46          & 88.52          \\ \cline{3-13} 
\multicolumn{2}{c|}{}                         & MaxEnt teacher    & 82.83          & 92.03          & 88.49          & 91.26          & 88.65          & 85.98          & 82.46          & 90.81          & 90.39          & 87.41          \\ \cline{3-13} 
\multicolumn{2}{c|}{}                         & w/o KD          & 74.44          & 90.78          & 85.42          & 84.83          & 83.87          & 77.98          & 78.52          & 83.73          & 83.19          & 80.86          \\  
\multicolumn{2}{c|}{}                         & token-level KD          & 75.08          & 90.95          & 85.88          & 85.10          & 84.25          & 78.40          & \textbf{79.52} & 84.92          & 85.50          & 82.09          \\  
\multicolumn{2}{c|}{}                         & Struct. KD     & \textbf{75.41}\rlap{$^{\dagger\ddagger}$} & \textbf{91.04}\rlap{$^{\dagger\ddagger}$} & \textbf{86.25}\rlap{$^{\dagger\ddagger}$} & \textbf{85.28}\rlap{$^{\dagger\ddagger}$} & \textbf{84.50}\rlap{$^{\dagger\ddagger}$} & \textbf{78.49}\rlap{$^{\dagger\ddagger}$} & 79.48\rlap{$^{\dagger}$}          & \textbf{85.28}\rlap{$^{\dagger\ddagger}$} & \textbf{85.66}\rlap{$^{\dagger\ddagger}$} & \textbf{82.23}\rlap{$^{\dagger\ddagger}$} \\ \hline
\multicolumn{2}{c|}{\multirow{3}{*}{\textbf{Case 4}}}  & Teacher           & 82.38          & 92.41          & 88.77          & 90.72          & 88.57          & 86.96          & 83.11          & 91.41          & 92.05          & 88.38          \\ \cline{3-13} 
\multicolumn{2}{c|}{}                         & w/o KD          & 74.44          & 90.78          & 85.42          & 84.83          & 83.87          & 77.98          & 78.52          & 83.73          & 83.19          & 80.86          \\  
\multicolumn{2}{c|}{}                         & Struct. KD     & \textbf{74.90}\rlap{$^{\dagger}$} & \textbf{91.21}\rlap{$^{\dagger}$} & \textbf{85.82}\rlap{$^{\dagger}$} & \textbf{85.20}\rlap{$^{\dagger}$} & \textbf{84.28}\rlap{$^{\dagger}$} & \textbf{78.66}\rlap{$^{\dagger}$} & \textbf{78.97}\rlap{$^{\dagger}$} & \textbf{83.83}\rlap{$^{\dagger}$} & \textbf{83.34}\rlap{$^{\dagger}$} & \textbf{81.45}\rlap{$^{\dagger}$} \\ 
\hlineB{4}
\end{tabular}%
\caption{Results of F1 scores for NER task on labeled datasets}
\label{tab:appendix_ner_labeled}
\end{table*}


\begin{table*}[t]
\centering

\begin{tabular}{c|l|l|c|ccccc}
\hlineB{4}
\multicolumn{3}{c|}{Dataset}                                              &                      & \multicolumn{5}{c}{WikiAnn with Unlabeled data}                                                   \\ \hline
\multicolumn{3}{c|}{Scenario}                                             & \# Unlabeled sent.   & de             & en             & es             & nl             & avg            \\ \hline\hline
\multicolumn{2}{c|}{\multirow{10}{*}{\textbf{Case 1a}}} & Teacher         &                      & 86.98          & 83.80          & 91.85          & 91.46          & 88.52          \\ \cline{3-9} 
\multicolumn{2}{c|}{}                                   & Top-1        & \multirow{3}{*}{3K}  & 80.66          & 79.85          & 87.79          & 88.44          & 84.19          \\ \cline{3-3} 
\multicolumn{2}{c|}{}                                   & Pos. KD + Top-1    &                      & 81.56          & \textbf{81.40} & 88.10          & 88.55          & 84.91          \\ \cline{3-3} 
\multicolumn{2}{c|}{}                                   & Struct. KD + Top-1    &                      & \textbf{81.88}\rlap{$^{\dagger\ddagger}$} & 81.23\rlap{$^{\dagger}$}          & \textbf{88.66}\rlap{$^{\dagger\ddagger}$} & \textbf{89.20}\rlap{$^{\dagger\ddagger}$} & \textbf{85.24}\rlap{$^{\dagger\ddagger}$} \\ \cline{3-9} 
\multicolumn{2}{c|}{}                                   & Top-1        & \multirow{3}{*}{10k} & 82.27          & 80.32          & 88.78          & 88.23          & 84.90          \\ \cline{3-3} 
\multicolumn{2}{c|}{}                                   & Pos. KD + Top-1   &                      & 82.01          & \textbf{81.53}          & 89.28          & 88.99          & 85.45          \\ \cline{3-3}  
\multicolumn{2}{c|}{}                                   & Struct. KD + Top-1    &                      & \textbf{82.34}\rlap{$^{\dagger\ddagger}$} & 81.27\rlap{$^{\dagger}$} & \textbf{89.85}\rlap{$^{\dagger\ddagger}$} & \textbf{89.19}\rlap{$^{\dagger\ddagger}$} & \textbf{85.66}\rlap{$^{\dagger\ddagger}$} \\ \cline{3-9} 
\multicolumn{2}{c|}{}                                   & Top-1        & \multirow{3}{*}{30k} & \textbf{84.20}          & 81.19          & 90.21          & 89.36          & 86.24          \\ \cline{3-3}  
\multicolumn{2}{c|}{}                                   & Pos. KD + Top-1     &                      & 84.12          & \textbf{82.56} & 89.82          & 89.53          & 86.51          \\ \cline{3-3} 
\multicolumn{2}{c|}{}                                   & Struct. KD + Top-1    &                      & 84.17\rlap{$^{\dagger\ddagger}$} & 82.14\rlap{$^{\dagger}$}          & \textbf{90.41}\rlap{$^{\dagger\ddagger}$} & \textbf{89.84}\rlap{$^{\dagger\ddagger}$}              & \textbf{86.64}\rlap{$^{\dagger\ddagger}$} \\ \hline
\multicolumn{2}{c|}{\multirow{10}{*}{\textbf{Case 2a}}} & Teacher         &                      & 86.98          & 83.80          & 91.85          & 91.46          & 88.52          \\ \cline{3-9} 
\multicolumn{2}{c|}{}                                   & Top-1        & \multirow{3}{*}{3K}  & 78.82          & 78.48          & 85.54          & 86.77          & 82.40          \\ \cline{3-3}  
\multicolumn{2}{c|}{}                                   & token-level KD + Top-1       &                      & \textbf{79.84} & 79.18          & 85.89          & 87.36          & 83.07          \\ \cline{3-3}  
\multicolumn{2}{c|}{}                                   & Struct. KD + Top-1    &                      & 79.82\rlap{$^{\dagger}$}          & \textbf{79.41}\rlap{$^{\dagger\ddagger}$} & \textbf{86.36}\rlap{$^{\dagger\ddagger}$} & \textbf{87.75}\rlap{$^{\dagger\ddagger}$} & \textbf{83.34}\rlap{$^{\dagger\ddagger}$} \\ \cline{3-9} 
\multicolumn{2}{c|}{}                                   & Top-1        & \multirow{3}{*}{10k} & 80.75          & 78.53          & 86.93          & 87.30          & 83.38          \\ \cline{3-3}  
\multicolumn{2}{c|}{}                                   & token-level KD + Top-1        &                      & 80.71          & 79.23          & \textbf{87.82} & 87.80          & 83.89          \\ \cline{3-3}  
\multicolumn{2}{c|}{}                                   & Struct. KD + Top-1    &                      & \textbf{81.07}\rlap{$^{\dagger\ddagger}$} & \textbf{79.41}\rlap{$^{\dagger\ddagger}$} & 87.77\rlap{$^{\dagger\ddagger}$}          & \textbf{87.99}\rlap{$^{\dagger\ddagger}$} & \textbf{84.06}\rlap{$^{\dagger\ddagger}$} \\ \cline{3-9} 
\multicolumn{2}{c|}{}                                   & Top-1        & \multirow{3}{*}{30k} & 82.49          & 79.43          & 88.78          & 88.74          & 84.86          \\ \cline{3-3}  
\multicolumn{2}{c|}{}                                   & token-level KD + Top-1        &                      & 82.35          & 80.42          & \textbf{89.32} & \textbf{88.84} & 85.23          \\ \cline{3-3}  
\multicolumn{2}{c|}{}                                   & Struct. KD + Top-1    &                      & \textbf{83.06}\rlap{$^{\dagger\ddagger}$} & \textbf{80.43}\rlap{$^{\dagger\ddagger}$} & 89.02\rlap{$^{\dagger}$}          & 88.62\rlap{$^{\dagger}$}          & \textbf{85.28}\rlap{$^{\dagger\ddagger}$} \\ \hline
\multicolumn{2}{c|}{\multirow{7}{*}{\textbf{Case 4}}}   & Teacher         &                      & 86.96          & 83.11          & 91.41          & 92.05          & 88.38          \\ \cline{3-9} 
\multicolumn{2}{c|}{}                                   & Top-1        & \multirow{2}{*}{3K}  & 78.41          & 77.22          & \textbf{85.82}          & 86.94          & 82.10          \\ \cline{3-3}  
\multicolumn{2}{c|}{}                                   & Struct. KD + Top-1    &                      & \textbf{78.80}\rlap{$^{\dagger}$} & \textbf{78.00}\rlap{$^{\dagger}$} & 85.75  & \textbf{87.22}\rlap{$^{\dagger}$} & \textbf{82.44}\rlap{$^{\dagger}$} \\ \cline{3-9}
\multicolumn{2}{c|}{}                                   & Top-1        & \multirow{2}{*}{10k} & 79.59          & 77.53          & 87.85 & \textbf{87.51}   & 83.12                 \\ \cline{3-3}  
\multicolumn{2}{c|}{}                                   & Struct. KD + Top-1    &                      & \textbf{80.04}\rlap{$^{\dagger}$} & \textbf{78.06}\rlap{$^{\dagger}$}  & \textbf{88.03}\rlap{$^{\dagger}$} & 87.40   & \textbf{83.38}\rlap{$^{\dagger}$}        \\ \cline{3-9} 
\multicolumn{2}{c|}{}                                   & Top-1        & \multirow{2}{*}{30k} & 81.47          & 78.59          & 89.46          & 88.80          & 84.58          \\ \cline{3-3}  
\multicolumn{2}{c|}{}                                   & Struct. KD + Top-1    &                      & \textbf{81.85}\rlap{$^{\dagger}$} & \textbf{79.57}\rlap{$^{\dagger}$} & \textbf{89.55}\rlap{$^{\dagger}$} & \textbf{89.13}\rlap{$^{\dagger}$} & \textbf{85.03}\rlap{$^{\dagger}$} \\ 
\hlineB{4}
\end{tabular}%
\caption{Results of F1 scores for NER task on unlabeled datasets}
\label{tab:appendix_ner_unlabel}
\end{table*}

\begin{table*}[t]
\centering

\begin{tabular}{l|c|ccccc}
\hlineB{4}
              & \multicolumn{1}{l|}{}                   & \multicolumn{5}{c}{WikiAnn}                                                       \\ \hline
\textbf{Case 3}        & \multicolumn{1}{l|}{\#Unlabeled sent.} & eu             & fa             & he             & ta             & Avg.           \\ \hline\hline
Teacher       & \multicolumn{1}{l|}{}                   & 67.92          & 40.30          & 58.68          & 57.14          & 56.01          \\ \hline
Top-1      & \multirow{3}{*}{3k}                     & 41.77          & 37.88          & 41.32          & 43.46          & 41.11          \\ \cline{1-1} 
Token-level KD + Top-1      &                                         & 52.67          & 26.32          & 36.22          & 38.45          & 38.42          \\ \cline{1-1} 
Struct. KD + Top-1 &                                         & \textbf{53.69}\rlap{$^{\dagger\ddagger}$} & \textbf{42.02}\rlap{$^{\dagger\ddagger}$} & \textbf{42.75}\rlap{$^{\dagger\ddagger}$} & \textbf{42.66}\rlap{$^{\dagger\ddagger}$} & \textbf{45.28}\rlap{$^{\dagger\ddagger}$} \\ \hline
Top-1      & \multirow{3}{*}{10k}                    & 58.63          & 34.65          & 43.37          & 57.18          & 48.46          \\ \cline{1-1}  
Token-level KD + Top-1      &                                         & 58.87          & 28.63          & 41.62          & 54.35          & 45.87          \\ \cline{1-1}  
Struct. KD + Top-1 &                                         & \textbf{62.50}\rlap{$^{\dagger\ddagger}$}  & \textbf{39.72}\rlap{$^{\dagger\ddagger}$}  & \textbf{46.22}\rlap{$^{\dagger\ddagger}$}  & \textbf{58.27}\rlap{$^{\dagger\ddagger}$} & \textbf{51.68}\rlap{$^{\dagger\ddagger}$} \\ \hline
Top-1      & \multirow{3}{*}{30k}                    & 74.37          & 35.70          & 55.12          & 63.78          & 57.24          \\ \cline{1-1} 
Token-level KD + Top-1     &                                         & 70.98          & 29.44          & 55.50          & 63.39          & 54.83          \\ \cline{1-1} 
Struct. KD + Top-1 &                                         & \textbf{75.66}\rlap{$^{\dagger\ddagger}$} & \textbf{38.08}\rlap{$^{\dagger\ddagger}$}  & \textbf{58.52}\rlap{$^{\dagger\ddagger}$}  & \textbf{64.69}\rlap{$^{\dagger\ddagger}$}  & \textbf{59.24}\rlap{$^{\dagger\ddagger}$} \\ 
\hlineB{4}
\end{tabular}%
\caption{Result of F1 scores of zero shot transfer experiment on NER task}
\label{tab:appendix_zeroshot}
\end{table*}

\subsection{Results of Parsing task}
Tabel \ref{tab:appendix_parsing_labeled} and \ref{tab:appendix_parsing_unlabeled} represent the results of experiments of Parsing. Our structural KD approaches significantly outperform the other approaches in all cases.
UAS and LAS in these tables were dependency parsing metrics, and they refer to unlabeled attachment score and labeled attachment score respectively.

\begin{table*}[t]
\centering

\begin{tabular}{c|l|c|c}
\hlineB{4}
 \multirow{2}{*}{metric} &               & \textbf{Case 1b} & \textbf{Case 2b} \\ \hhline{|~|~|-|-|}
                &               & PTB              & PTB              \\ \hline\hline
\multirow{3}{*}{UAS}   & Teacher       & 95.96            & 96.04            \\ \cline{2-4} 
                      & w/o KD      & 91.78            & 91.78            \\ \cline{2-2} 
                      & Struct. KD & \textbf{93.56}\rlap{$^{\dagger}$}   & \textbf{93.56}\rlap{$^{\dagger}$}   \\ \hline
\multirow{3}{*}{LAS}   & Teacher       & 94.24            & 94.29            \\ \cline{2-4} 
                      & w/o KD      & 89.85            & 89.85            \\ \cline{2-2} 
                      & Struct. KD & \textbf{91.83}\rlap{$^{\dagger}$}   & \textbf{91.78}\rlap{$^{\dagger}$}   \\ 
\hlineB{4}
\end{tabular}
\caption{Result of F1 scores of Parsing task with labeled dataset. Note that all our approaches are significantly stronger than the baseline.}
\label{tab:appendix_parsing_labeled}
\end{table*}

\begin{table*}[t]
\small
\centering
\setlength\tabcolsep{4pt}
\begin{tabular}{l|l|l|l|r|r|r|r|l|l|l|l|l|l}
\hlineB{4}
\multicolumn{2}{l|}{}            & \multicolumn{6}{c|}{\textbf{Case 1b}}                                                                                                         & \multicolumn{6}{c}{\textbf{Case 2b}}                                                               \\ \hline
\multicolumn{2}{c|}{}            & \multicolumn{6}{c|}{PTB with Unlabeled data}                                                                                                              & \multicolumn{6}{c}{PTB with Unlabeled data}                                                                    \\ \hline
Metric               &            & 3k             & 10k            & \multicolumn{1}{l|}{30k} & \multicolumn{1}{l|}{50k} & \multicolumn{1}{l|}{100k} & \multicolumn{1}{l|}{Avg.} & 3k             & 10k            & 30k            & 50k            & 100k           & Avg.           \\ \hline\hline
\multirow{2}{*}{UAS} & Top-1   & 92.00 & 92.52 & 93.22                    & 93.69                    & 94.25                     & 93.14                     & 91.99          & 92.44          & 93.16          & 93.69          & 94.26          & 93.11          \\ \cline{2-2} 
                     & Struct. KD + Top-1 & \textbf{93.71}\rlap{$^{\dagger}$}& \textbf{93.93}\rlap{$^{\dagger}$} & \textbf{94.26}\rlap{$^{\dagger}$}           & \textbf{94.58}\rlap{$^{\dagger}$}           & \textbf{94.84}\rlap{$^{\dagger}$}            & \textbf{94.26}\rlap{$^{\dagger}$}            & \textbf{93.67}\rlap{$^{\dagger}$} & \textbf{93.90}\rlap{$^{\dagger}$} & \textbf{94.30}\rlap{$^{\dagger}$} & \textbf{94.64}\rlap{$^{\dagger}$} & \textbf{94.89}\rlap{$^{\dagger}$} & \textbf{94.28}\rlap{$^{\dagger}$} \\ \hline
\multirow{2}{*}{LAS} & Top-1   & 90.03 & 90.62 & 91.44                    & 91.99                    & 92.61                     & 91.34                     & 90.03          & 90.59          & 91.41          & 91.98          & 92.66          & 91.33          \\ \cline{2-2} 
                     & Struct. KD + Top-1 & \textbf{91.98}\rlap{$^{\dagger}$} & \textbf{92.24}\rlap{$^{\dagger}$} & \textbf{92.63}\rlap{$^{\dagger}$}         & \textbf{93.00}\rlap{$^{\dagger}$}          & \textbf{93.28}\rlap{$^{\dagger}$}            & \textbf{92.63}\rlap{$^{\dagger}$}            & \textbf{91.94}\rlap{$^{\dagger}$} & \textbf{92.18}\rlap{$^{\dagger}$} & \textbf{92.66}\rlap{$^{\dagger}$} & \textbf{93.04}\rlap{$^{\dagger}$} & \textbf{93.31}\rlap{$^{\dagger}$} & \textbf{92.63}\rlap{$^{\dagger}$} \\ 
\hlineB{4}
\end{tabular}
\caption{The accuracy of Parsing task with unlabeled dataset (in thousand). Note that all our approaches are significantly stronger than the baseline.}
\label{tab:appendix_parsing_unlabeled}
\end{table*}



\end{document}